%% file: main.tex
\documentclass{article}

\usepackage{microtype}
\usepackage{graphicx}
\usepackage{booktabs} 

\usepackage{times}
\usepackage{epsfig}
\usepackage{amsmath}
\usepackage{amssymb}
\usepackage{dblfloatfix}
\usepackage{paralist}
\usepackage{subcaption}

\usepackage{dsfont}

\usepackage{style/mysymbols}

\usepackage[normalem]{ulem}

\usepackage{hyperref}

\usepackage[algo2e]{algorithm2e}

\newcommand{\change}[1] {\textcolor[rgb]{0,0,0}{#1}}

\usepackage[accepted]{style/icml2019}

\icmltitlerunning{Counterfactual Visual Explanations}

\input{style/space_saver.tex}
\begin{document}
\graphicspath{figures/}

\twocolumn[
\icmltitle{Counterfactual Visual Explanations}



\icmlsetsymbol{equal}{*}

\begin{icmlauthorlist}
\icmlauthor{Yash Goyal}{gt}
\icmlauthor{Ziyan Wu}{sie}
\icmlauthor{Jan Ernst}{sie}
\icmlauthor{Dhruv Batra}{gt}
\icmlauthor{Devi Parikh}{gt}
\icmlauthor{Stefan Lee}{gt}
\end{icmlauthorlist}

\icmlaffiliation{gt}{Georgia Institute of Technology}
\icmlaffiliation{sie}{Siemens Corporation}

\icmlcorrespondingauthor{Yash Goyal}{ygoyal@gatech.edu}

\icmlkeywords{Machine Learning, ICML, Interpretability, Explanations, Counterfactual}

\vskip 0.3in
]



\printAffiliationsAndNotice{}  

\begin{abstract}


In this work, we develop a technique to produce \emph{counterfactual visual explanations}. 
Given a `query' image $I$ for which a vision system predicts class $c$, 
a counterfactual visual explanation identifies how 
$I$ could change such that the system would output a different specified class $c'$. 
To do this, we select a `distractor' image $I'$ that the system predicts as class $c'$ and 
identify spatial regions in $I$ and $I'$ such that replacing the identified region 
in $I$ with the identified region in $I'$ would push the system towards classifying $I$ as $c'$. 
We apply our approach to multiple image classification datasets 
generating qualitative results showcasing the interpretability and discriminativeness of our counterfactual explanations.
To explore the effectiveness of our explanations in teaching humans, we present machine teaching experiments for the task of fine-grained bird classification.
We find that users trained to distinguish bird species fare better when given access to counterfactual 
explanations in addition to training examples.

\end{abstract}

\input{sections/intro_v3.tex}
\input{sections/approach.tex}

\input{sections/related_work.tex}
\input{sections/experiments.tex}

\input{sections/teaching.tex}
\input{sections/conclusion.tex}


\section*{Appendix}
In this appendix, we provide additional results on the SHAPES~\cite{NMN} dataset in Appendix I.
\input{sections/supp.tex}

\section*{Acknowledgements}

We thank Ramprasaath Selvaraju for useful discussions.
This work was supported in part by NSF, AFRL, DARPA, ECASE-Army, Siemens, Samsung, Google, Amazon, ONR YIPs and ONR Grants N00014-16-1-\{2713,2793\}. The views and conclusions contained herein are those of the authors and should not be interpreted as necessarily representing the official policies or endorsements, either expressed or implied, of the U.S. Government, or any sponsor.




\end{document}

%% file: style/space_saver.tex
\belowdisplayskip \abovedisplayskip

\newlength{\sectionReduceTop}
\newlength{\sectionReduceBot}
\newlength{\subsectionReduceTop}
\newlength{\subsectionReduceBot}
\newlength{\abstractReduceTop}
\newlength{\abstractReduceBot}
\newlength{\captionReduceTop}
\newlength{\captionReduceBot}
\newlength{\subsubsectionReduceTop}
\newlength{\subsubsectionReduceBot}

\newlength{\eqnReduceTop}
\newlength{\eqnReduceBot}

\newlength{\horSkip}
\newlength{\verSkip}

\newlength{\figureHeight}

%% file: sections/intro_v3.tex
\section{Introduction}

When we ask for an explanation of a decision, either implicitly or explicitly we do so expecting the answer to be given with respect to likely alternatives or specific unselected outcomes -- \emph{``For situation X, why was the outcome Y and not Z?''}
A common and useful technique for providing such discriminative explanations is through counterfactuals -- i.e.~describing what changes to the situation would have resulted in arriving at the alternative decision -- \emph{``If X was X*, then the outcome would have been Z rather than Y.''}
As computer vision systems achieve increasingly widespread and consequential applications, the need to explain their decisions in arbitrary circumstances is growing as well -- for example, questions of safety \emph{``Why did the self-driving car misidentify the fire hydrant as a stop sign?''} or fairness \emph{``Why did the traveler surveillance system select John Doe for additional screening?''} will need answers.

\begin{figure}[t]
\centering
  \includegraphics[width=0.95\linewidth, clip=true, trim=0px 7px 0px 0px]{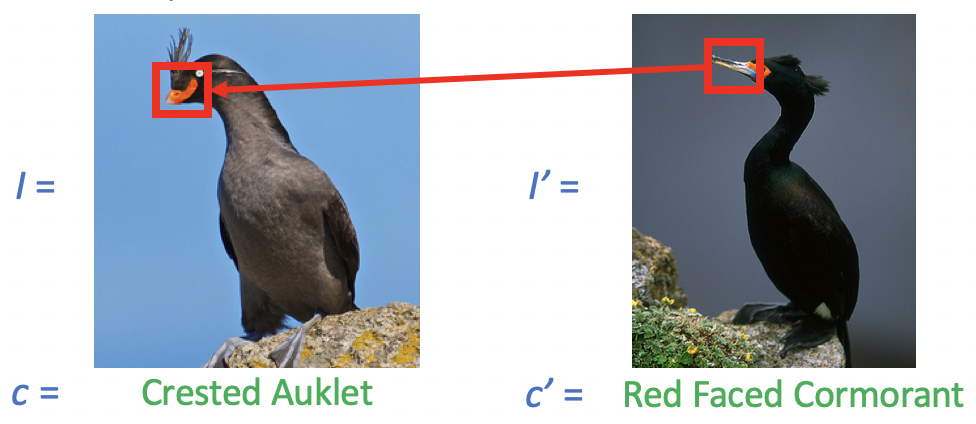}\\[-8pt]
  \caption{Our approach generates counterfactual visual explanations for a query image $I$ (left) -- explaining why the example image was classified as class $c$ (\emph{Crested Auklet}) rather than class $c'$ (\emph{Red Faced Cormorant}) by finding a region in a distractor image $I'$ (right) and a region in the query $I$ (highlighted in red boxes) such that if the highlighted region in the left image looked like the highlighted region in the right image, the resulting image $I^*$ would be classified more confidently as $c'$.}
  \vspace{-10pt}
\label{fig:teaser}
\end{figure}

While deep learning models have shown unprecedented (and occasionally super-human) capabilities in a range of computer vision tasks \cite{ImageNet, LFWTech}, they achieve this level of performance at the cost of becoming increasingly inscrutable compared to simpler models. However, understanding the decisions of these deep models is important to guide practitioners to design better models, evaluate fairness, establish appropriate trust in end-users, and to enable machine-teaching for tasks where these models have eclipsed human performance. 

As these systems are increasingly been deployed in real world applications, interpretability of machine learning systems (particularly deep learning models) has become an active area of research \cite{simonyan_arxiv13,GuidedBackprop,lime_sigkdd16,SanityNIPS2018,DoshiKim2017Interpretability}. Many of these works have focused on identifying regions in an input image that most contributed to the final model decision \cite{GuidedBackprop,lime_sigkdd16,simonyan_arxiv13,gradcam_arxiv,zhang2016top}  -- i.e.~producing explanations via feature attribution. However, these approaches do not consider alternative decisions or identify hypothetical adjustments to the input which could result in different outcomes -- i.e.~they are neither discriminative nor counterfactual.

In this work, we study how such \textbf{counterfactual visual explanations} can be generated to explain the decisions of deep computer vision systems by identifying what and how regions of an input image would need to change in order for the system to produce a specified output. Consider the example in Figure \ref{fig:teaser}, a computer vision system identifies the left bird as \emph{Crested Auklet}. A standard feature attribution explanation approach may identify the bird's crown, slender neck, or colored beak as important regions on this image. However, when considering an alternative such as the \emph{Red Faced Cormorant} (shown right) many of these key regions would be shared across both birds. In contrast, our approach provides a counterfactual explanation by identifying the beak region in both images -- indicating that if the bird on the left had a similar beak to that on the right, then the system would have output \emph{Red Faced Cormorant}.

More concretely, given a query image $I$ for which the system predicts class $c$, we would like to generate a faithful, counterfactual explanation which identifies how $I$ could change such that the system would output a specified class $c'$. To do this, we select a distractor image $I'$ which the system predicts as class $c'$ and identify spatial regions in $I$ and $I'$ such that replacing the identified region in $I$ with the identified region in $I'$ would push the system towards classifying $I$ as $c'$. We formalize this problem and present greedy relaxations that sequentially execute such region edits.

We apply our approach to SHAPES~\cite{NMN}, MNIST~\cite{lecun1998gradient}, Omniglot~\cite{omniglot} and the Caltech-UCSD Birds (CUB) 2011~\cite{CUB_200_2011} datasets generating qualitative results showcasing the intepretability and discriminativeness of our counterfactual explanations.
Our design of SHAPES dataset also enables us to quantitatively evaluate our generated explanations.
The simplistic nature of MNIST and Omniglot images allows us to generate an imagination of how the query image would ``look'' like if the discriminative region in the query image is replaced by the discriminative region in the distractor image.
For CUB, we present an analysis of our counterfactual explanations utilizing segmentation and keypoint annotations present in the dataset which shows that our explanations highlight discriminative parts of the birds.

In addition to being more discriminative and interpretable, we think this explanation modality is also compelling from a pedagogical perspective, an important and relatively less studied application for interpretability.
Good teachers explain why something is a particular object and why it's not some other object. 
Similarly, these counterfactual explanations can be useful in the context of machine teaching, \ie AI teaching humans, especially for tasks where AI systems outperform untrained humans.

We apply our approach to a fine-grained bird classification task in which deep models perform significantly better than untrained humans on the Caltech-UCSD Birds (CUB) 2011 dataset~\cite{CUB_200_2011}.
We hypothesize that our counterfactual visual explanations from a deep model trained for this task can help in teaching humans where in the image to look at (\eg neck, beak, wings, \etc of the bird) in order to identify the correct bird category. 
For example, for the birds shown in Figure \ref{fig:teaser}, 
most people not specifically trained in bird recognition would not know the difference between these two birds. 
But, given a counterfactual visual explanation from our approach ``If the highlighted region in the left image looked like the highlighted region in the right image, the left image would look more like a \textit{Red Faced Cormorant}'', an untrained human is more likely to learn the differences between these two birds as compared to only showing example images for both birds.

To explore the effectiveness of our explanations in teaching humans the fine-grained differences between birds, we designed a human study where we train and test humans for this task of bird classification.
Through our human studies, we found that users trained to discern between bird species fare better when given access to counterfactual explanations in addition to training examples.

In summary, we make the following contributions: we
\begin{compactenum}
    \item propose an approach to generate counterfactual visual explanations, \ie what region in the image made the model predict class $c$ instead of class $c'$?
    \item show that our counterfactual explanations from deep models can help in teaching humans via human studies. 
\end{compactenum}

%% file: sections/approach.tex
\section{Approach}

\begin{figure*}[b]
\centering
\includegraphics[width=1.85\columnwidth, clip=true, trim=0px 0px 0px 5px]{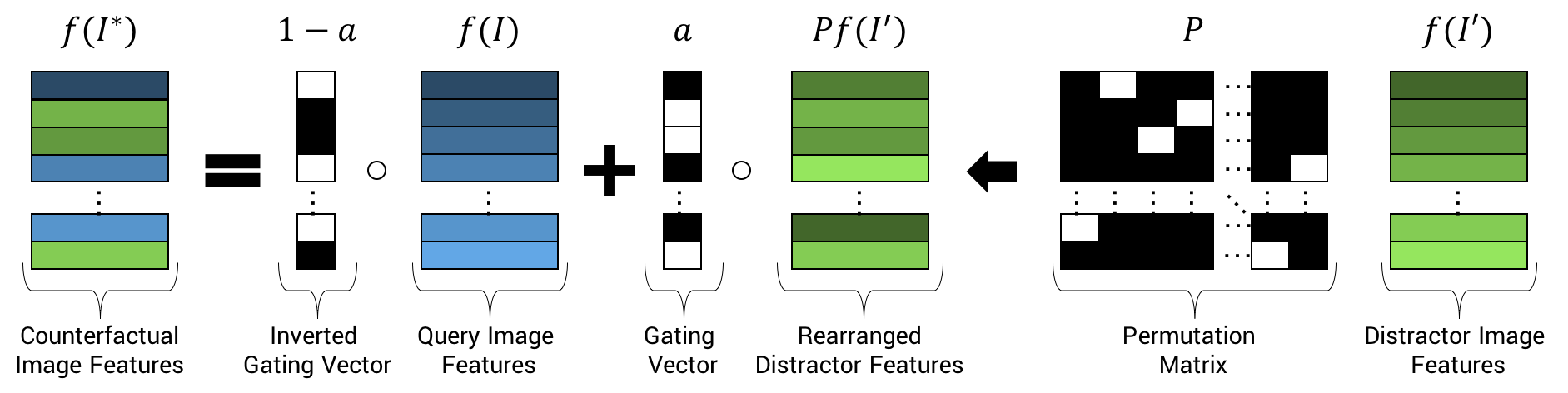}\\[-12pt]
\caption{To parameterize our counterfactual explanations, we define a transformation that replaces regions in the query image $I$ with those from a distractor $I'$. Distractor image features $f(I')$ are first rearranged with a permutation matrix $P$ and then selectively replace entries in $f(I)$ according to a binary gating vector $a$. This allows arbitrary spatial cells in $f(I')$ to replace arbitrary cells in $f(I)$.}
\label{fig:transform}
\vspace{-8pt}
\end{figure*}

In order to explain a query image $I$ relative to a distractor $I'$ under some trained network, we seek to identify the key discriminative regions in both the images such that replacing these regions in $I'$ with those in $I$ would lead the network to change its decision about the query to match that of the distractor. In \secref{sec:minedit} we formalize this problem and then present two greedy solutions: exhaustive search (\secref{sec:search}) and a continuous relaxation (\secref{sec:relax}).

\begin{figure}[t]
\centering
\includegraphics[width=0.95\columnwidth]{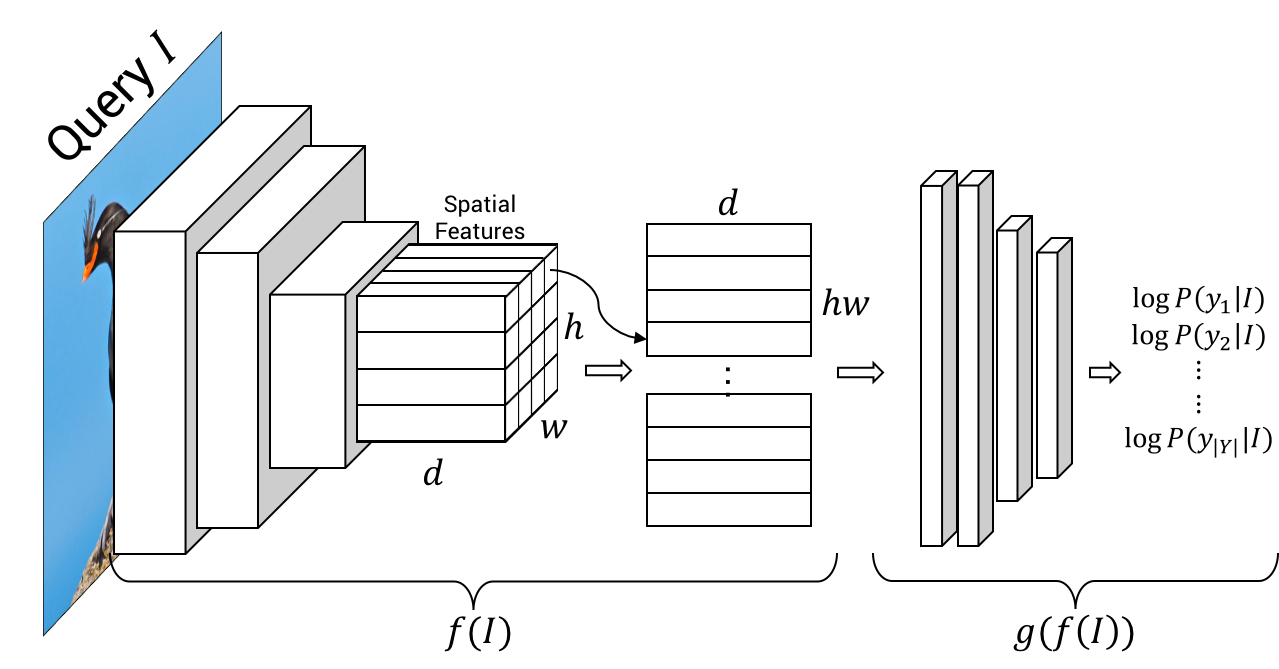}\\[-13pt]
\caption{We decompose a CNN as a spatial feature extractor $f(I)$ and a decision network $g(f(I))$ as shown above.}
\label{fig:decomp}
\vspace{-12pt}
\end{figure}

\subsection{Minimum-Edit Counterfactual Problem}
\label{sec:minedit}
Consider a deep convolutional network taking as input an image $I\in \mathcal{I}$ and predicting log-probability output $\log P(\mathbf{Y}|I)$ over the output space $\mathbf{Y}$. 
For the sake of this discussion and without loss of generality, we will consider a decomposition of the network into two functional components -- a spatial feature extractor and a decision network -- as shown in Figure \ref{fig:decomp}. First, $f: \mathcal{I} \rightarrow \mathbb{R}^{h*w\times d}$ maps the image to a $h\times w\times d$ dimensional spatial feature which we reshape to a $hw \times d$ matrix, where $h$ and $w$ are the spatial dimensions and $d$ is the feature size (\ie number of channels). Second, $g: \mathbb{R}^{hw\times d} \rightarrow \mathbb{R}^{|\mathbf{Y}|}$ takes this feature and predicts log-probabilities over output classes $\mathbf{Y}$. We can then write the network as a whole as $\log P(\mathbf{Y}|I) = g(f(I))$. For notational convenience, we let $g_{c}(f(I))$ denote the log-probability of class $c$ for image $I$ under this network.

Given a query image $I$ for which the network predicts class $c$, we would like to produce a 
counterfactual explanation which identifies how $I$ could change such that the network would 
output a specified distractor class $c'$. However, the space of possible changes to $I$ is 
immense and directly optimizing pixels in $I$ to maximize $\log P(c'|I)$ is unlikely to yield 
interpretable results \cite{tyka_2015}. Instead, we consider changes towards a distractor image 
$I'$ which the network already predicts as class $c'$. 

Given these two images, we would like to make a transformation $T$ from $I$ to $I^* = T(I,I')$ 
such that $I^*$ appears to be an instance of class $c'$ to the trained model $g(f(\cdot))$. One natural
way of perform this transformation is by replacing regions in image $I$ with regions in image $I'$.
However at the extreme, we could simply set $I^*=I'$ and replace $I$ entirely. To avoid such trivial
solutions while still providing meaningful change, we would like to apply a minimality constraint 
on the number of transferred regions. Hence, our approach tries to find the minimum number of region replacements from $I'$ to $I$ to generate $I^*$ such that the trained model classifies $I^*$ as an instance of class $c'$.
We call this the \emph{minimum-edit counterfactual problem}. Rather than considering the actual image regions themselves, we consider the spatial feature maps $f(I), f(I') \in \mathbb{R}^{hw\times d}$ corresponding to image regions.

We formalize this transformation as depicted in Figure \ref{fig:transform}. Let $P \in \mathbb{R}^{hw \times hw}$ be a permutation matrix that rearranges the spatial cells of $f(I')$ to align with spatial cells of $f(I)$ and let $a\in \mathbb{R}^{hw}$ be a binary vector indicating whether to replace each spatial feature in image $I$ with spatial features from image $I'$ (value of 1) or to preserve the features of $I$ (value of 0). 
We can then write the transformation from $I$ to $I^*$ in spatial feature space $f(^*)$ as
\begin{equation}
    f(I^*) = (\mathds{1} - \mathbf{a}) \circ f(I) + \mathbf{a} \circ Pf(I')
\label{eq:fI*}
\end{equation}
where $\mathds{1}$ is a vector of all ones and $\circ$ represents the Hadamard product between a vector and a matrix obtained by broadcasting the vector to match the matrix's dimensions and then taking the Hadamard product between the broadcasted vector and the matrix. Note that as $\mathbf{a}$ is a binary vector, minimizing its norm corresponds to minimizing the number of edits from $I'$ to $I$.

With this notation in hand, we can write the 
 \emph{minimum-edit counterfactual problem}, \ie
 minimizing the number of edits to transform $I$ to $I^*$ such that the predicted class for the transformed image features $f(I^*)$ as defined in Eq. \ref{eq:fI*} is the distractor class $c'$, 
as the following:
\begin{equation}
\begin{aligned}
    & \underset{P, \mathbf{a}}{\text{minimize~~~}}
      ||a||_1 && \\
    & ~~~~~~\text{s.t.~~~~~~}
     c' = \mbox{argmax~~} g( (\mathds{1}-\mathbf{a}) \circ f(I) + \mathbf{a} \circ Pf(I') ) \\
    & ~~~~~~~~~~~~~~~~a_i \in \{0,1\}~~\forall i \mbox{~~and~~} P \in \calP  
\end{aligned}%
\label{eq:hard_opt}
\end{equation}
\noindent where $\calP$ is the set of all $hw{\times}hw$ permutation matrices. Given the resulting $\mathbf{a}$ and $P$, we can extract the set of pairs of spatial cells involved in the edits as $\mathcal{S} = \{(i,j,i',j')\}) \mid a_{i*j} = 1 \land P_{{i*j},{i'*j'}}=1\}$.

After optimization, the resulting vector $\mathbf{a}$ provides the discriminative attention map on image $I$ indicating which spatial cells in $I$ were edited with features copied from $I'$ and $P_{i^*}$, where $i^*$ = $\underset{i}{\text{argmax}}$ $a_i$, provides the discriminative attention map on the distractor image $I'$ indicating which source cells those features were taken from. 

Solving this problem directly is quite challenging -- requiring identifying the minimum subset of $hw*hw$ possible edits that changes the model's decision. To put this in perspective, there are $O((h*w)^{2+k})$ such subsets of size $k$ \ie $k$ cells in $I$ being replaced by  $k$ cells in $I'$. Even for modest feature sizes of $h{=}w{=}16$, this quickly scales over a million candidates for $k=2$.

In the following sections, we present two greedy sequential relaxations -- first, an exhaustive search approach keeping $\mathbf{a}$ and $P$ binary and second, a continuous relaxation of $\mathbf{a}$ and $P$ that replaces search with an optimization.

\begin{figure}[t]
    \centering
    \includegraphics[clip=true, trim=5px 0px 0px 0px, width=0.85\linewidth]{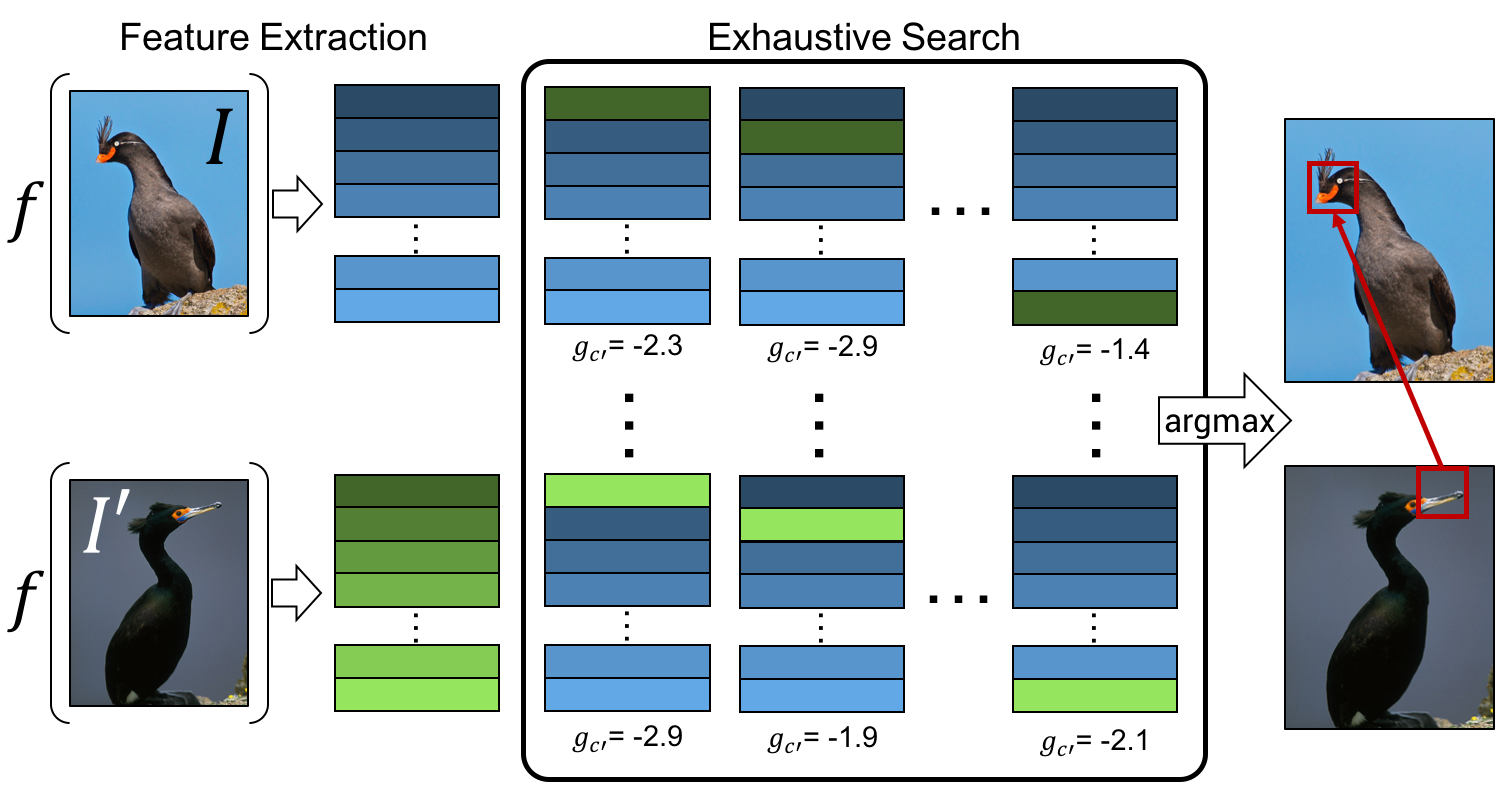}\\[-12pt]
    \caption{In our exhaustive best-edit search, we check all pairs of query-distractor spatial locations and select whichever pair maximizes the log probability of the distractor class $c'$.}
\label{fig:search}
\end{figure}

\subsection{Greedy Sequential Exhaustive Search}
\label{sec:search}

Rather than solve Eq.~\ref{eq:hard_opt}, we consider greedily making edits to $I$ until the cumulative effect changes the model's decision. That is to say, we sequentially find single edits that maximizes the gain in output log-probability $g_{c'}(\cdot)$ for $c'$. 

We can write this \emph{best-edit} sub-problem as
\begin{equation}
\begin{aligned}
    & \underset{P, a}{\text{maximize}}
    & & g_{c'}( (\mathds{1}-\mathbf{a}) \circ f(I) + \mathbf{a} \circ Pf(I') ) \\
    & ~~~~~~\text{s.t.}
    & & ||\mathbf{a}||_1 = 1,~~a_i \in \{0,1\}~~\forall i\\
    & & & P\in{\calP}
\end{aligned}
\label{eq:one_opt}
\end{equation}
where $\mathbf{a}$ is a binary vector and $P$ is a permutation matrix as before. Rather than minimizing $||\mathbf{a}||_1$ as in Eq.~\ref{eq:hard_opt}, we instead constrain it to be one-hot -- indicating the edit in $I$ which maximizes the model log-probability $g_{c'}$.

One straight-forward approach to solving Eq.~\ref{eq:one_opt} is through exhaustive search -- that is to say, evaluating $g_{c'}$ after replacing features for each of the $h*w\times h*w$ cells in $f(I)$ with those of each of the cells in $f(I')$. 
As shown in Fig.~\ref{fig:search}, one step of our exhaustive search approach consists of first replacing the features in one spatial cell of $f(I)$ by features of one spatial cell in $f(I')$, and then passing these modified convolutional features through the the rest of the classifier network $g(.)$ to compute the log-probability of the distractor class $c'$.
We then repeat this procedure for all permutations of cell locations in $f(I)$ and $f(I')$. 
The pair of cells that result in the greatest log-probability for the distractor class $c'$ are the most discriminative spatial cells in $f(I)$ and $f(I')$.

In order to approximately minimize the objective in Eq.~\ref{eq:hard_opt} \ie the number of edits, we run this search sequentially multiple times (excluding previously selected edits) until the predicted class changes from $c$ to $c'$, i.e. $g_{c'} > g_c$. We outline this procedure in Algorithm \ref{alg:search}. 

This procedure requires evaluating $g(\cdot)$ $O(h^2w^2k)$ times where $k$ is the average number of edits before the decision changes. In the next subsection, we provide a continuous relaxation of the best-edit problem amicable to gradient based solutions -- resulting in fewer evaluation calls on average.

\begin{figure}[t]
  \centering
  \resizebox{0.9\columnwidth}{!}{
  \begin{minipage}{\linewidth}
    \begin{algorithm}[H]
 \KwData{query image $I$ with class $c$, distractor $I'$ with class $c'$}
 \KwResult{list of edits $\mathcal{S}$ that change the model decision }
 
$\mathcal{S} \leftarrow [~~]$ ~~~ $F^* \leftarrow f(I)$ ~~~ $F' \leftarrow f(I')$\;

\;

\tcc{Until decision is changed to $c'$}
\While{$~c' \neq argmax~~g(F^*)~$}{
    \;\;
    \tcc{Find single best edit excluding previously edited cells in $S$}\;
    
    $i,j' \leftarrow \mbox{BestEdit}(F^*, F', S)$\;
    
    \;\;
    
    \tcc{Apply the edit and record it}\;
    
    $F^*_{i,*} = F'_{j',*}$\;
    
    $\mathcal{S}.append(\{i,j'\})$\;
    
 }
 \caption{Greedy Sequential Search}
 \label{alg:search}
\end{algorithm}
  \end{minipage}}
\end{figure}

\subsection{Continuous Relaxation}
\label{sec:relax}
We formulate the best-edit problem defined in Eq.~\ref{eq:one_opt} as a tractable optimization problem by relaxing the constraints.

First, we loosen the restriction that $\mathbf{a}$ be binary -- allowing it to instead be a point on the simplex (i.e. non-negative and summing to one) corresponding to a distribution over which cells in $f(I)$ to edit. Second, we allow a similar softening of the constraints on $P$, restricting it to be a right stochastic matrix (i.e. non-negative with rows $\mathbf{p_i}^T$ summing to one) corresponding to distributions over cells in $f(I')$ to be copied from. We write this relaxed objective as:
\begin{equation}
\begin{aligned}
    & \underset{P, a}{\text{maximize}} &
    & g_{c'}( (\mathds{1}-\mathbf{a}) \circ f(I) + \mathbf{a} \circ Pf(I') ) \\
    & \text{~~~~~~s.t.}
    & & ||a||_1 = 1,~~ a_i \geq 0~~\forall i\\
    & & & ||p_{i}||_1 = 1 \quad \forall i, ~~ P_{i,j} \geq 0~~\forall i,j
\end{aligned}
\label{eq:soft_one}
\end{equation}
To always satisfy the constraints in Eq.~\ref{eq:soft_one}, we reparameterize $\mathbf{a}$ and $P$ in terms of auxilliary variables $\mathbf{\alpha}$ and $M$ respectively. Specifically, we define $\mathbf{a}{=}\sigma(\alpha)$ and $\mathbf{p}_{i}^T{=}\sigma(\mathbf{m_{i}}^T)$ where $\sigma(\cdot)$ is the softmax function: $a_i = \frac{e^{\alpha_i}}{\sum_j e^{\alpha_j}}$. In this way, the non-negativity and unit norm constraints on $\mathbf{a}$ and $P$ are ensured while we are free to optimize $\alpha$ and $M$ unconstrained via gradient descent. We use gradient descent with a learning rate of $0.3$.

In this soft version, cells in $f(I')$ can be copied to more
than one location or copied fractionally through non-binary
entries in $P$ or $\mathbf{a}$; however, 
by applying entropy losses on $\mathbf{a}$ and rows of $P$ (minimizing their entropy),
we can recover a nearly binary solution for $\mathbf{a}$ and the rows of $P$.

We apply this approach as the \emph{best-edit} search procedure in lieu of exhaustive search presented in Section \ref{sec:search} -- iteratively selecting the best-edit until the decision changes.

To summarize our approach, we defined a formulation of minimum-edit counterfactual problem and introduced two greedy sequential relaxations -- an exhaustive search approach in Sec.~\ref{sec:search} and a continuous relaxation in Sec.~\ref{sec:relax}.

%% file: sections/related_work.tex
\section{Related Work}

\textbf{Visual Explanations.}
\change{
Various feature attribution methods have been proposed in the recent years which highlight ``important'' regions in the input image which led the model to make its prediction. 
Many of these approaches are gradient based \cite{simonyan_arxiv13,GuidedBackprop,gradcam_arxiv}, using backpropagation-like techniques and upsampling to generate visual explanations.
Another type of feature attribution methods are reference based approaches \cite{ruth_vedaldi_iccv17, dabkowski_gal_nips17, zintgraf2016visualizing, dhurandhar2018explanations, chang2018explaining}, which focus on the change in classifier outputs with respect to perturbed input images \ie input images where parts of the image have been masked and replaced with various references such as mean pixel values, blurred image regions, random noise, outputs of generative models, etc. 
In similar spirit, a concurrent work \citet{chang2018explaining} use a trained generative model to fill in the masked image regions from the unmasked regions.
More relevant to our approach, \citet{dhurandhar2018explanations} find minimal regions in the input image which should be necessarily present/absent for a particular classification.
But, all the above works focus on generating visual explanations that highlight regions in an image which made the model predict a class $c$.
On the other hand, we focus on a more specific task of generating counterfactual visual explanations that highlight what and how
regions of an image would need to change in order for the model to predict a distractor class $c'$ instead of the predicted class $c$. 
}

\textbf{Counterfactual Explanations.}
To tackle a similar task as ours, \citet{lisa_grounding_expl} learn a model to generate a counterfactual explanation for why a model predicted a class $c$ instead of class $c'$.
But our approach is different from theirs in 3 significant ways: 1) their explanation is in natural language while our explanation is visual, 2) their approach requires additional attribute annotations while ours doesn't, and 3) their explanation is the output of a separate learned model (raising concerns regarding its faithfulness to the target model's prediction) while our explanation is directly generated from the target model based on the receptive field of the model's neurons and, hence, is faithful by design.

\textbf{Machine Teaching.}
Machine teaching \cite{zhu_machineteaching} works have mainly focused on how to show examples to humans so that they can learn a task better.
Many of these approaches focus on selecting or ordering examples to be shown to humans in order to maximize their information gain.
As deep learning models achieve superhuman performance in some tasks, it is natural to ask if they can in turn act as instructors to help improve human ability.
To our knowledge, only one other work \cite{explainteachcvpr18} uses visual explanations for machine teaching. They use saliency map explanations generated from \citet{zhou2016cvpr} along with heuristics to select good examples to be shown to human learners. To compare with an equivalent setting of this work to ours, we ran a baseline human study with explanations generated from GradCAM \cite{gradcam_arxiv}, which has been shown to generate more discriminative visual explanations as compared to \citet{zhou2016cvpr}.

%% file: sections/experiments.tex
\section{Experiments}
\label{sec:experiments}

We apply our approach on four different datasets -- SHAPES \cite{NMN} (in supplement due to space constraints), MNIST \cite{lecun1998gradient}, Omniglot~\cite{omniglot} and Caltech-UCSD Birds (CUB) 2011 Dataset \cite{CUB_200_2011}, and present results showcasing the intepretability and discriminativeness of our counterfactual explanations.

\textbf{Common Experimental Settings}
In all our experiments, we operate in output space of the last convolutional layer in the CNN but our approach is equally applicable to the output of any convolutional layer. Further, all qualitative results shown are with the exhaustive search approach presented in Section \ref{sec:search} as we are operating on relatively small images.  

In our experiments on CUB (Sec.~\ref{sec:cub}), we find the continuous relaxation presented in Sec.~\ref{sec:relax} achieves identical solutions to exhaustive search for 79.98\% of instances and on average achieves distractor class probability that is 92\% of the optimal found via exhaustive search -- suggesting its usefulness for larger feature spaces.


\subsection{MNIST}
\label{sec:mnist}

We begin in the simple setting of hand-written digit recognition on the MNIST dataset \cite{lecun1998gradient}. This setting allows us to explore our counterfactual approach in a domain well-understood by humans.

We train a CNN model consisting of 2 convolutional layers and 2 fully-connected layers on this dataset. This network achieves 98.4\% test accuracy -- note this is well below state-of-the-art but this is not important for our purposes.
Under this model, the size of spatial features is $4 \times 4 \times 20$.

\begin{figure}[t]
\centering
  \includegraphics[ width=0.8\linewidth, clip=true, trim=40px 320px 40px 20px]{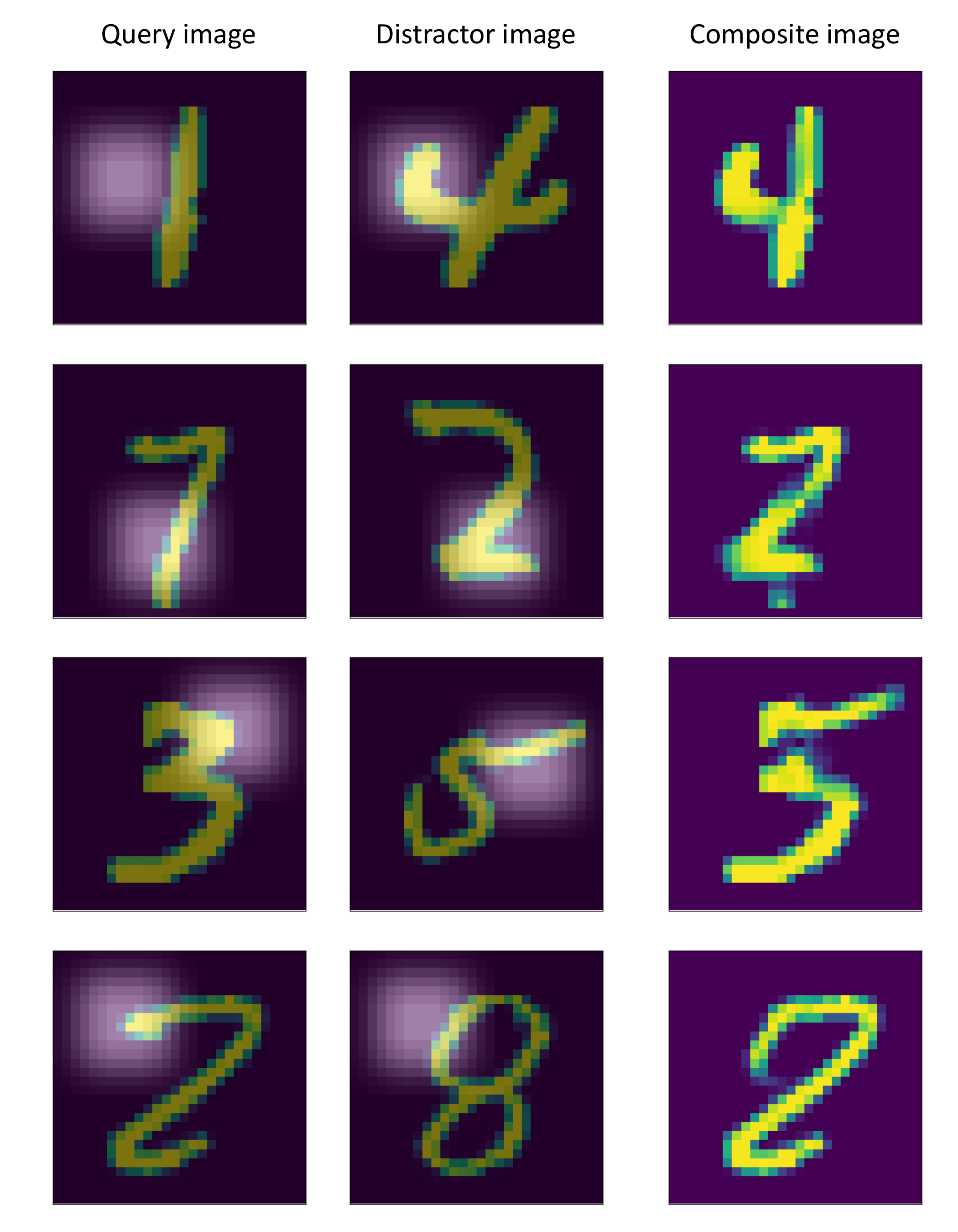}\\[-8pt]
  \caption{Results on MNIST \cite{lecun1998gradient} dataset. The first two columns show the query and distractor images, each with their identified discriminative region highlighted. The third column shows composite images created by making the corresponding replacement in pixel space.} 
\label{fig:mnist_results}
\vspace{-10pt}
\end{figure}

\textbf{Qualitative Results.}
We examine counterfactual explanations for randomly selected distractor class $c'$ and corresponding image $I'$. Sample results are shown in \figref{fig:mnist_results}. 

\begin{figure}
\centering
\includegraphics[clip=true, trim=0px 20px 0px 200px, width=0.9\columnwidth]{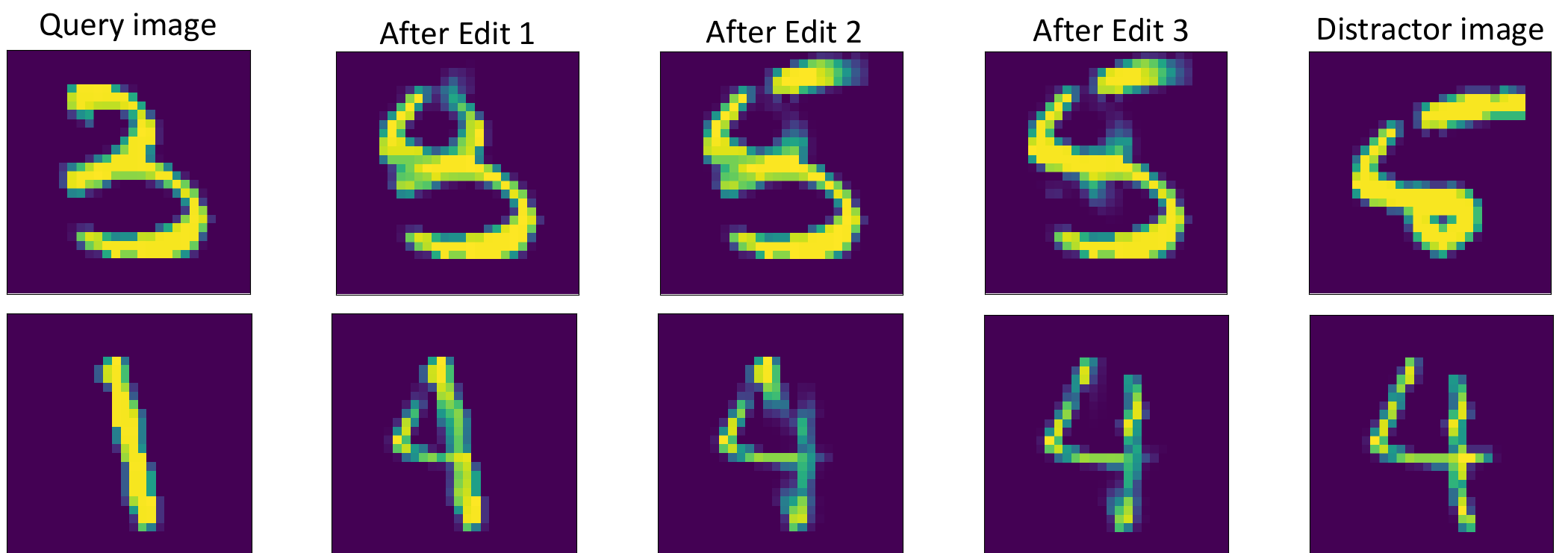}\\[-10pt]
\caption{Examples of multiple edits on MNIST digits.}
\label{fig:multiedit}
\vspace{-16pt}
\end{figure}

The first two columns show the query and distractor images and highlight the \emph{best-edit} regions. We produce this highlight based on the receptive field of the convolutional feature selected through our approach.
The third column depicts a composite image generated in pixel space by aligning and superimposing the highlighted region centers.
We note that our approach operates in the convolutional feature space and we present this composite as visualization.

For the first example (first row), our approach finds that that if the left side stroke (the highlighted region) in the distractor image (2nd column) replaced the highlighted region (background) in the query image, the query image would be more likely to belong to the distractor class `4'. As we can see in the composite, the resulting digit does appear to be a `4'. As a reminder, the network was not trained to consider such transformation, rather our approach is identifying the key discriminative edits.
In the third example, our approach finds that if the upper curve of the `3' in the query image instead looked like the horizontal stroke of the `5' in the distractor image, the query image would be more likely to belong to the distractor class `5'.

\textbf{Quantitative Analysis.}
On average, it takes our approach 2.67 edits to change the model's prediction from $c$ to $c'$. Examples with multiple (3) edits are shown in \figref{fig:multiedit}. As edits are taken greedily, often the first edit makes the most significant change (second row); however, for complex transformation like $3 \rightarrow 5$ (top row) multiple edits are needed.
\change{Our approach takes 15 $\mu$s per image on a Titan XP GPU.}


\subsection{Omniglot}
\label{sec:omniglot}

\begin{figure}[t]
\centering
  \includegraphics[clip=true, trim=40px 60px 40px 20px, width=0.8\linewidth]{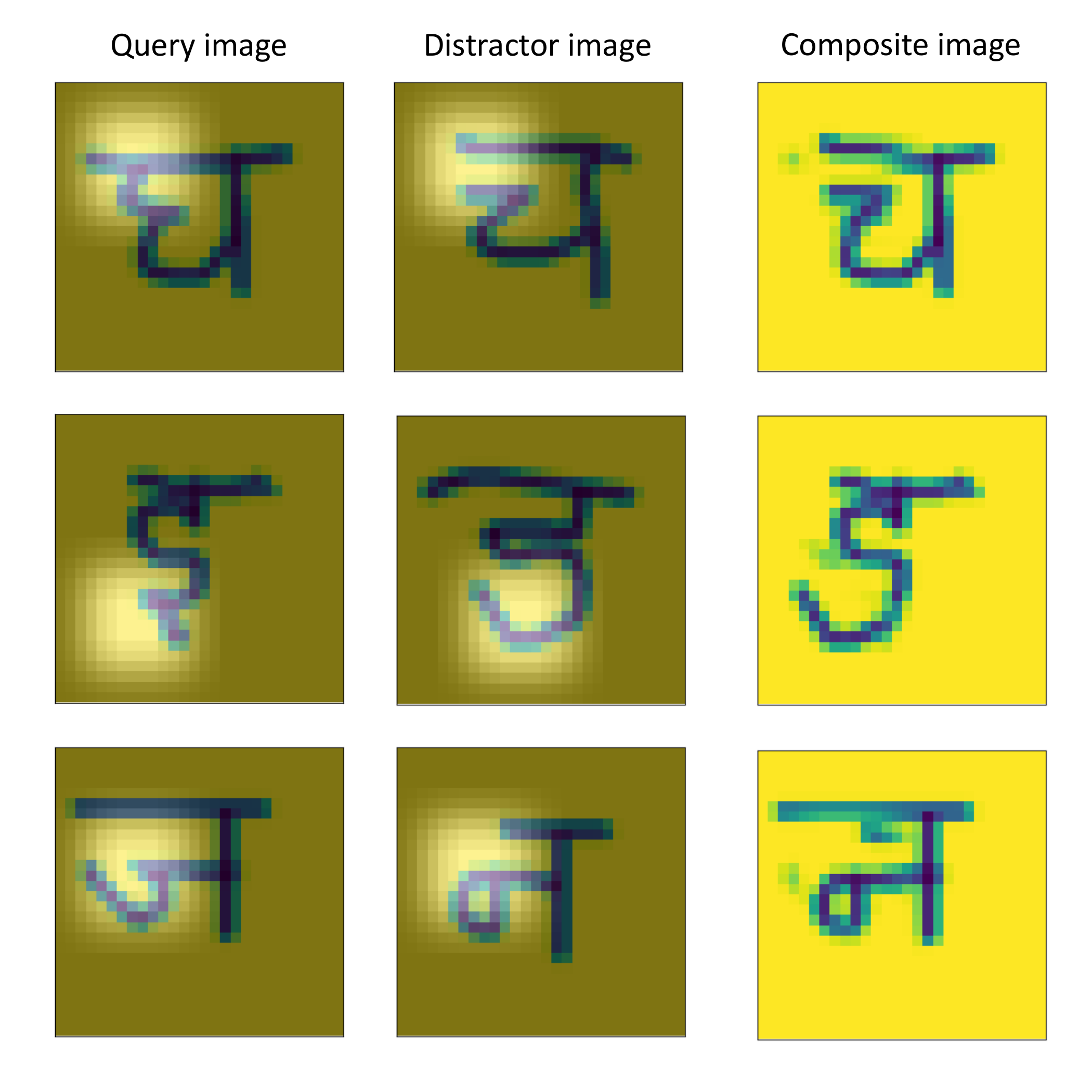}\\[-10pt]
  \caption{Qualitative results on the Omniglot dataset.}
\label{fig:omniglot_results}
\vspace{-20pt}
\end{figure}

We move on to the Omniglot dataset \cite{omniglot} containing images of hand-written characters from 50 writing systems. Like MNIST, these images are composed of simple pen strokes; however, most humans are not going to a priori know the difference between characters.
Hence, Omniglot is an ideal ``mid-way point'' between our MNIST and CUB experiments. 

We experiment with the `Sanskrit' writing system consisting of 42 characters with 20 images each. We created a random train/test split of 80/20\% to train the classification model. We use the same architecture as in the MNIST experiments, resizing the Omniglot images to match. This network achieves 66.8\% test accuracy.

\textbf{Qualitative Results.}
As before, we examine single-edit counterfactual explanations for randomly selected distractors. Qualitative results are shown in \figref{fig:omniglot_results}. Our approach finds appropriate counterfactual edits to shift the character towards the distractor even given their complex shape.

\textbf{Quantitative Analysis.}
On average, it takes our approach 1.46 edits to change the model's prediction from $c$ to $c'$. \change{Runtime is 9 $\mu$s per image on a Titan XP GPU.}

\begin{figure}[t]
\centering
  \includegraphics[clip=true,trim=20px 20px 20px 20px, width=0.9\columnwidth]{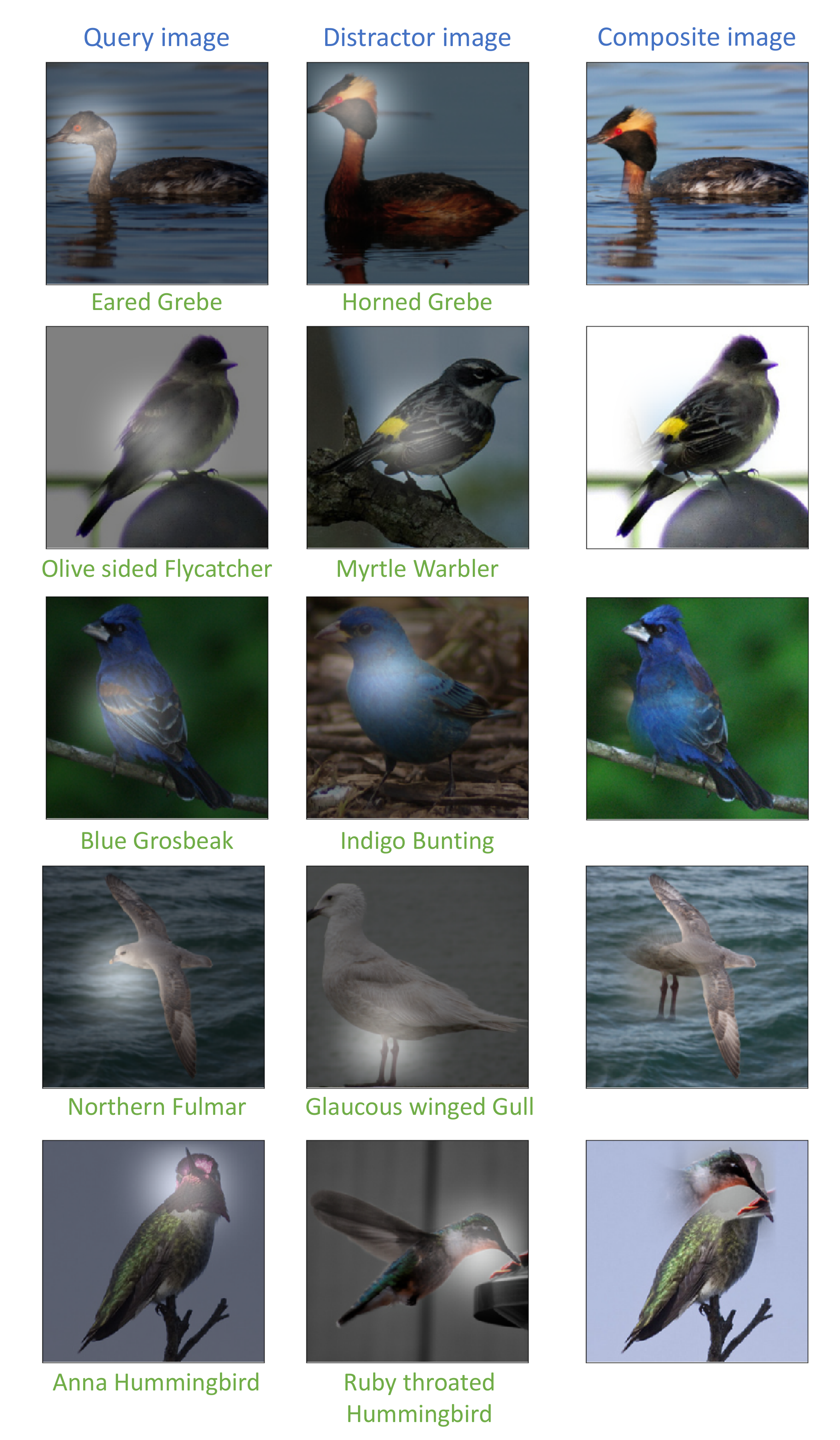}\\[-8pt]
  \caption{Qualitative results on CUB. Our counterfactual explanation approach highlights important attributes in the birds such as head plumage, yellow wing spots and texture on the wings.}
\label{fig:cub_results}
\vspace{-14pt}
\end{figure}

\begin{figure}[t]
\centering
  \includegraphics[clip=true, width=0.86\columnwidth]{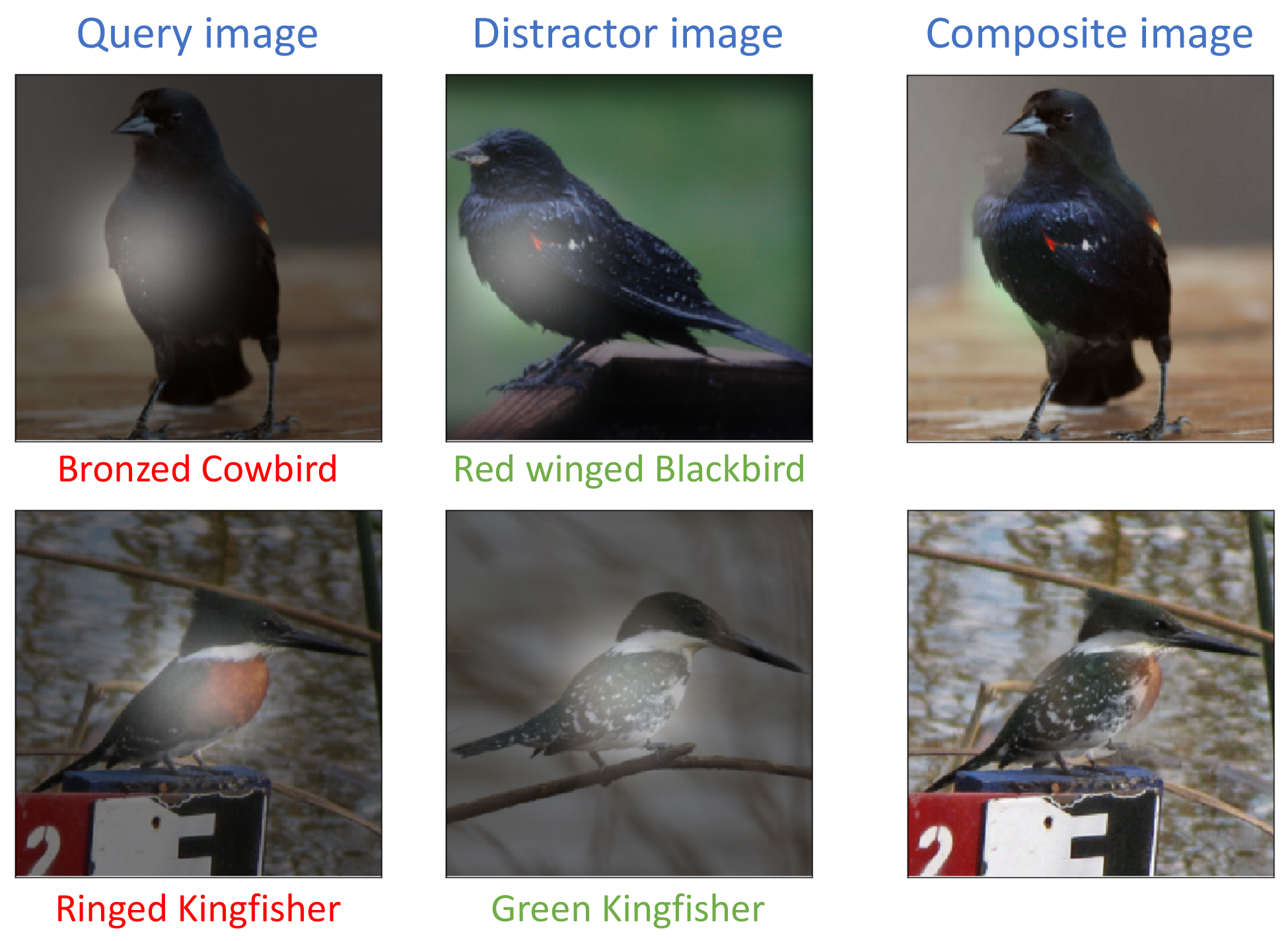}\\[-8pt]
  \caption{Qualitative results where the model's predictions are incorrect. Our counterfactual explanations with respect to the correct class highlights important attributes of the correct class which are not clearly visible in the query images such as red wing spot and texture on the wings.}
\label{fig:incorrect_pred_cub_results}
\vspace{-14pt}
\end{figure}


\subsection{Caltech-UCSD Birds (CUB)}
\label{sec:cub} 

Finally, we apply our approach to the Caltech-UCSD Birds (CUB) 2011 dataset \cite{CUB_200_2011} consisting of 200 bird species. It is one of the most commonly used datasets for fine-grained image classification and can be challenging for non-expert humans. Consequentially, we use this dataset for our machine teaching experiments.

We trained a VGG-16 \cite{Simonyan15} model on this dataset, which achieves 79.4\% test accuracy. 
The size of this feature space is 7 x 7 x 512.

Given an image $I$ with the predicted class $c$, we consider two ways of choosing the distractor class $c'$ -- random classes different from $c$ and nearest neighbor classes of $c$ in terms of average attribute annotations (provided with the dataset). 
The latter helps us in creating pairs of images ($I$, $I'$) which are very similar looking to each other. 
We sample the distractor image $I'$ with the predicted class $c'$ in two ways -- random images and nearest neighbor images of $I$ in terms of keypoint locations of the bird (provided with the dataset).
The latter helps us in creating pairs of images ($I$, $I'$) in which the birds are in similar orientation to each other.

\textbf{Qualitative Results.}
As before, single-edit qualitative results are shown in \figref{fig:cub_results}. The first three examples depict success cases where our approach identifies important features like head plumage, yellow wing spots, and wing coloration. In these cases, the simple composite visualization is quite telling. However, the bottom two rows show less interpretable results. In the fourth row, the query bird's head is replaced by the long legs of the distractor bird -- perhaps in an attempt to turn the bird shape around as shown in the composite. The fifth row seems to correctly identify the need to recolor the neck of the query bird; however, the composite looks poor due to pose misalignment.

\change{
Our explanations can also be helpful in debugging a model's mistakes. Such examples are shown in \figref{fig:incorrect_pred_cub_results}. 
In the first example, the query image is incorrectly classified as \textit{Bronzed Cowbird} instead of \textit{Red winged Blackbird} probably because the distinct feature of the correct class (a red spot on the wing) is not clearly visible. When an explanation is generated with respect to an image of the correct class, our approach copies over the red spot in order to increase the score of the correct class.
Similarly, in the second example, our approach highlights the distinct texture on the wings which is not clearly visible in the query image.
}

\textbf{Quantitative Analysis.} On average, it takes our approach 7.4 edits to change the model's prediction from $c$ to $c'$ if $c'$ is a random distractor class while it takes on average 5.3 edits if $c'$ is a nearest neighbor in terms of attributes.
\change{Runtimes are 1.85 and 1.34 sec/image for random and NN distractor classes respectively on a Titan XP GPU.}

To check the degree of dependence of our explanations on the choice of the distractor image, we compute ``agreement'' in the most discriminative spatial cell locations \ie outputs of the best-edit subproblem in image $I$ for different distractor images with the same predicted class $c'$.
This agreement is 78\% (a high value) implying that our approach highlights similar regions in image $I$ for different choices of distractor image $I'$ from class $c'$. Similarly, we compute ``agreement'' in the most discriminative spatial cell locations in image $I$ for different distractor classes $c'$. This agreement is 42\% (a low value) implying that our explanations on image $I$ differ based on the choice of the distractor class.

The CUB dataset also comes with dense annotations of bird regions and parts which we use to further analyze our approach. First, we compute how often our discriminative regions lie inside the bird segmentations and find that regions in both the images lie inside the bird segmentation 97\% of the times.

Further, we compute how often our discriminative regions lie near the important keypoints of birds such as neck, crown, wings, legs, \etc provided with the dataset. 
After running our explanation procedure until the model decision changes, we find they are near the keypoints of the bird 75\% of the times in the query image $I$ and 80\% of the times in the distractor image $I'$.
Our discriminative regions also highlight the same keypoint in both the birds 20\% of the times. This shows that our approach often replaces semantically meaningful keypoints between birds -- indicating that the underlying CNN has likely picked up on these keypoints without explicit supervision.

%% file: sections/teaching.tex
\section{Machine Teaching}
\label{sec:teaching}

\begin{figure}[t]
    \centering
    \resizebox{1\columnwidth}{!}{
    \begin{subfigure}[t]{0.46\columnwidth}
     \includegraphics[width=\textwidth, clip=true, trim=0pt 0pt 0pt 10pt]{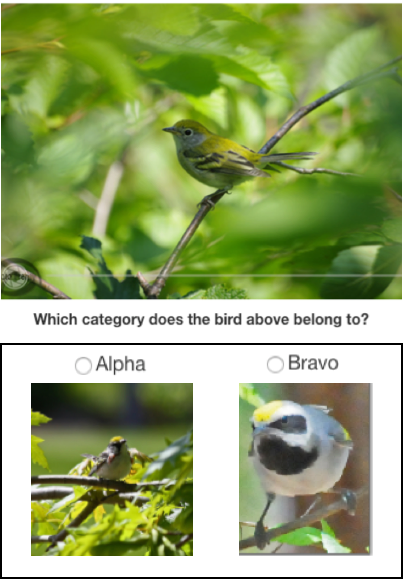}
     \caption{Training Interface}
     \label{fig:training}
    \end{subfigure}\hfill%
    \begin{subfigure}[t]{0.83\columnwidth}
     \includegraphics[width=\textwidth]{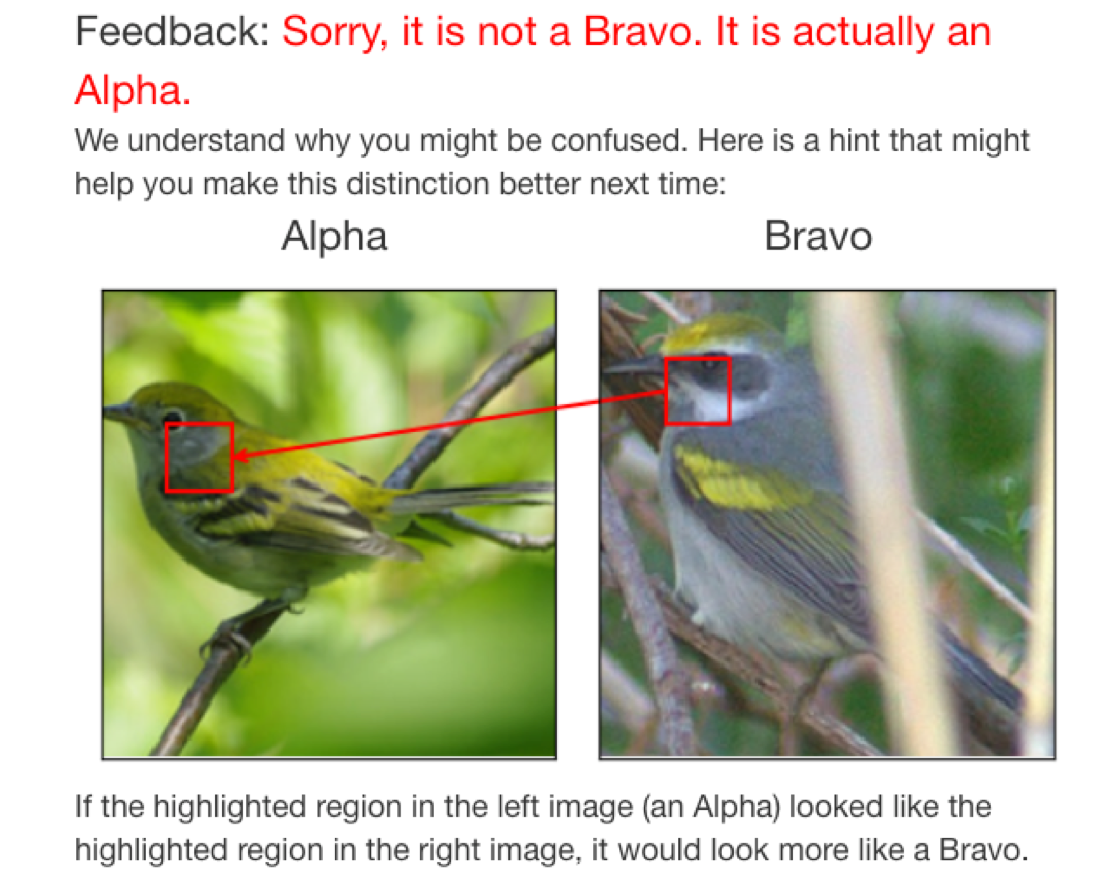}
     \caption{Feedback}
     \label{fig:feedback}
    \end{subfigure}\hfill%
    \begin{subfigure}[t]{0.62\columnwidth}
     \includegraphics[width=\textwidth]{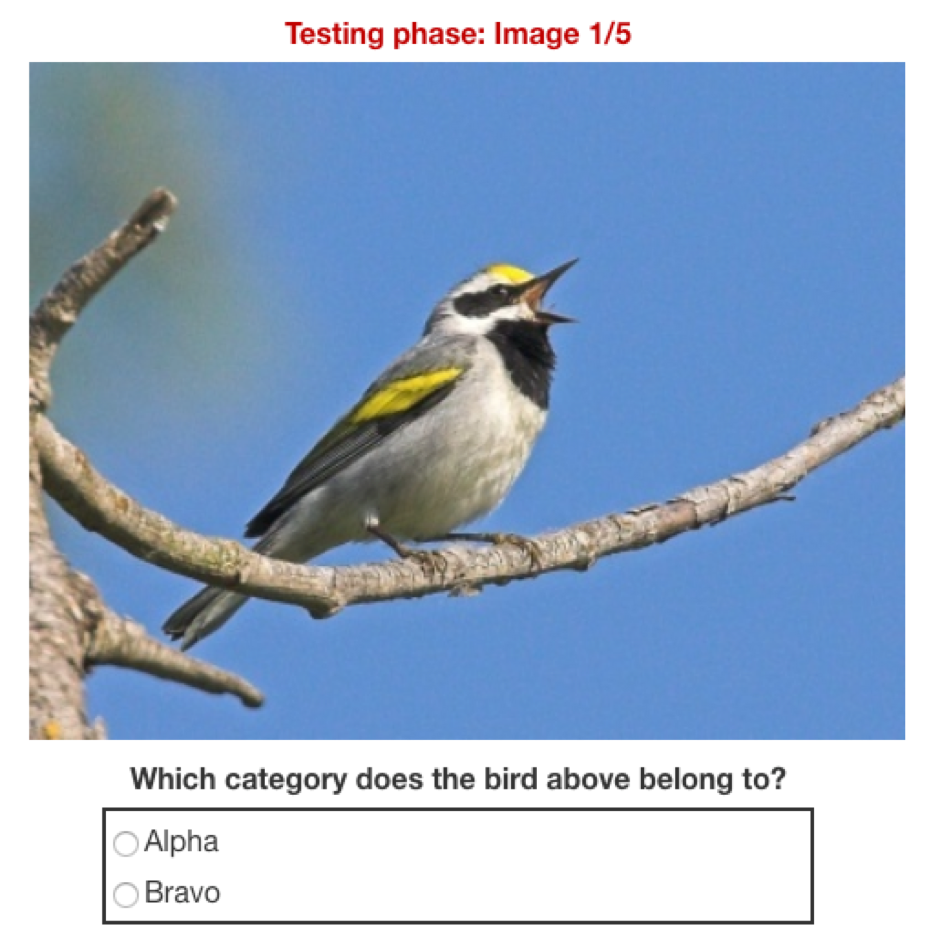}
     \caption{Testing Interface}
     \label{fig:testing}
    \end{subfigure}}\\[-5pt]
  \caption{Our machine teaching interface. During training phase (shown in (a)), if the participants choose an incorrect class, they are shown a feedback (shown in (b)) highlighting the fine-grained differences between the two classes. At test time (shown in (c)), they must classify the birds from memory.}
\label{fig:amt}
\vspace{-10pt}
\end{figure}

We apply our approach on CUB 2011 dataset \cite{CUB_200_2011} to guide humans where to look to distinguish bird categories. 
Towards this goal, we built a machine teaching interface for human subjects shown in Figure \ref{fig:amt}.

Since it is a hard task for humans, the interface consists of two phases-- Training and Testing.
In both the phases, we show the subjects an image and 2 bird category options denoted Alpha and Bravo, and ask them to assign the shown bird images to one of these categories.
During the Training phase, each of the category option is accompanied by an example image for subjects to compare against the presented image. 
Our training interface is shown in \figref{fig:training}.

If they choose an incorrect category, we show them some feedback. The feedback shows the discriminative attention maps generated by our approach using an image of the distractor class (Alpha if the correct class is Bravo and vice-versa) closest to the query image in keypoint space as the distractor image. The feedback in our interface is shown in \figref{fig:feedback}.

After the training phase, we test the human subjects.
During the testing phase, they are shown an image and the bird category options, but without the example images.
As in training, they have to select the option they think best fits the query test image. No feedback is given during testing.
Our test interface is shown in \figref{fig:testing}.

We compare this human study with two baselines which only differ in terms of the feedback shown to human subjects.
In the first baseline case, the feedback only conveys the information that the subject chose the incorrect category.
In the second and harder baseline, in place of our counterfactual explanations, the feedback shows a non-counterfactual, feature-attribution explanation generated via GradCAM \cite{gradcam_arxiv} highlighting the region in the image most salient for the predicted category.

\change{
The studies were conducted in smaller sessions. Each session consisted of teaching a participant 2 bird classes with 9 training and 5 test examples.
We conducted all studies with graduate students studying Machine Learning (ML) because of the level of difficulty of the task.
Total 26 graduate students participated, each participant (on average) worked on 6.2 sessions and spent 3 min 52 sec per session.} 

The mean test accuracy with counterfactual explanations is 78.77\%, with GradCAM explanations, it is 74.29\% while the mean test accuracy without any explanations is 71.09\%.
We find that our approach is statistically significantly better than the no-explanation baseline at a 87\% confidence level, and than the GradCam baseline at 51\% -- implying the need for further study to assess if counterfactual explanations improve machine teaching compared to feature attribution approaches.
\change{To examine the effect of a participant's familiarity to ML on their performance, we conducted a small study with 9 participants with no knowledge of ML, each working on 5 sessions. The test accuracy is 61.7\% without explanations and 72.4\% with our counterfactual explanations, trends consistent with the previous human study.}
Overall, this shows that counterfactual explanations from deep models can help teach humans by pointing them to appropriate parts to identify the correct bird category.

%% file: sections/conclusion.tex
\section{Conclusion}
\label{sec:conclusion}

In this work, we present an approach to generate counterfactual visual explanations -- answering the question ``How should the image $I$ be different for the model
to predict class \textit{c'} instead?''
We show our approach produces informative explanations for multiple datasets. Through a machine teaching task on fine-grained bird classification, we show that these  explanations can provide guidance to humans to help them perform better on this classification task.

%% file: sections/supp.tex

\section{Experiments on SHAPES}
\label{sec:shapes}
\textbf{Dataset.} 
To first evaluate our model on a simple setting, we created a dataset of SHAPES images \cite{NMN} for classification using the code released by the authors. 
This dataset consists of 3x3 grid images of size 30 pixels x 30 pixels. Only one out of the 9 cells contains a shape which can be either a circle, a square or a triangle, which is also the label of the image. 
Any of these shapes can take any of the three colors -- blue, green and red.
There is some small random perturbation in the size of each shape and in the pixel values of each color.

\textbf{Classification model.} We trained a simple CNN
consisting of 1 convolutional layer followed by 2 fully connected layers with 3 output classes.
The network achieves 100\% test accuracy, which is unsurprising due to the simplicity of the task. 

\textbf{Experimental settings.} 
For this task, the size of spatial features is 3 x 3 x 100.
We randomly choose a distractor class $c'$ different from the predicted class $c$, and a distractor image $I'$ from the set of images for which the model predicts $c'$.

\begin{figure}[]
\centering
  \includegraphics[width=\linewidth]{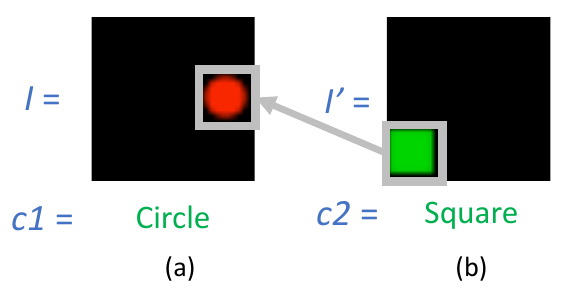}\\[-12pt] 
  \caption{Results on SHAPES images. Each image is made up of 3x3 cells, one of which contains a shape. (a) Our approach highlights the middle right cell in the image $I$ containing the \textit{circle} shape which led the model to predict the class \textit{Circle} instead of class \textit{Square}.
(b) In addition, our approach also highlights the bottom left cell containing the \textit{square} shape in image $I'$ of the distractor class \textit{Square} such that if the middle right cell in image $I$ looked like the bottom left cell in image $I'$, the model’s prediction would have been \textit{Square}.}
\label{fig:shapes}
\end{figure}

\textbf{Results.} 
Since these images are generated automatically, the cell location containing the shape is known for each image.
Hence, the correct discriminative attention maps are known for each pair of ($I$, $I'$) and the results of our approach can be quantitatively evaluated automatically.
We found that approach is able to find the accurate attention maps 100\% of the times.
An example is shown in Fig. \ref{fig:shapes}.